\documentclass[conference]{IEEEtran}
\IEEEoverridecommandlockouts
\usepackage{cite}
\usepackage{amsmath,amssymb,amsfonts}
\usepackage{algorithmic}
\usepackage{graphicx}
\usepackage{textcomp}
\usepackage{xcolor}
\usepackage{url}
\def\BibTeX{{\rm B\kern-.05em{\sc i\kern-.025em b}\kern-.08em
    T\kern-.1667em\lower.7ex\hbox{E}\kern-.125emX}}
\begin{document}


\title{Diffusion Graph Neural Networks and Dataset for Robust Olfactory Navigation in Hazard Robotics}

\author{\IEEEauthorblockN{Kordel K. France}
\IEEEauthorblockA{\textit{Dept. of Computer Science} \\
\textit{University of Texas at Dallas}\\
Richardson, TX, USA \\
kordel.france@utdallas.edu}
\and
\IEEEauthorblockN{Ovidiu Daescu}
\IEEEauthorblockA{\textit{Dept. of Computer Science} \\
\textit{University of Texas at Dallas}\\
Richardson, TX, USA \\
ovidiu.daescu@utdallas.edu}
}

\pagestyle{plain}

\maketitle

\begin{abstract}
Navigation by scent is a capability in robotic systems that is rising in demand. 
However, current methods often suffer from ambiguities, particularly when robots misattribute odours to incorrect objects due to limitations in olfactory datasets and sensor resolutions. 
To address challenges in olfactory navigation, we introduce a multimodal olfaction dataset along with a novel machine learning method using diffusion-based molecular generation that can be used by itself or with automated olfactory dataset construction pipelines.
This generative process of our diffusion model expands the chemical space beyond the limitations of both current olfactory datasets and training methods, enabling the identification of potential odourant molecules not previously documented.
The generated molecules can then be more accurately validated using advanced olfactory sensors, enabling them to detect more compounds and inform better hardware design. 
By integrating visual analysis, language processing, and molecular generation, our framework enhances the ability of olfaction-vision models on robots to accurately associate odours with their correct sources, thereby improving navigation and decision-making through better sensor selection for a target compound in critical applications such as explosives detection, narcotics screening, and search and rescue.
Our methodology represents a foundational advancement in the field of artificial olfaction, offering a scalable solution to challenges posed by limited olfactory data and sensor ambiguities\footnote{Code, models, and data are made available to the community at:
\newline
\url{https://huggingface.co/datasets/kordelfrance/olfaction-vision-language-dataset}
}.
\end{abstract}

\section{Introduction}
The human sense of smell is a complex and nuanced sensory modality, capable of distinguishing an extensive array of odourants. 
In recent years, artificial olfactory systems have been developed to detect sources of explosives, illegal drugs, and even perform forensic studies.
These systems, such as electronic noses, utilize sensor arrays and pattern recognition algorithms to detect and classify compounds \cite{Chowdhury2025_survey, covington2021_artificialolfactionsurvey21stcentury, gutierrez_osuna_machine_olfaction_survey_2002, kim_olfactory_sensor_survey_2022}.

\begin{figure}[t!]
  \centering
  \includegraphics[width=1.0\linewidth]{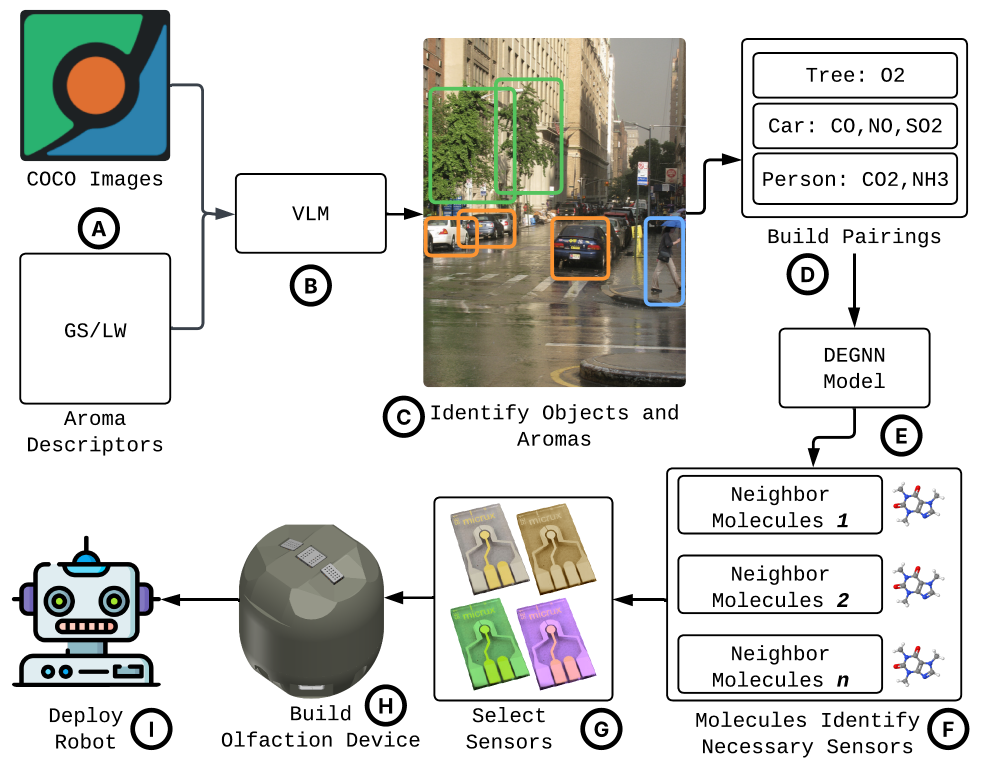}
  \caption{An example of how our dataset (A-D) and diffusion equivariant graph neural network (DEGNN) model (E-F) helps in olfactory robotics.  (A) Images are taken from the COCO dataset and aroma descriptors are taken from the GoodScents and Leffingwell Datasets. (B) The VLM is instructed to identify the objects (from the COCO classes) and the aromas in the images. (C) Labels of identified objects are returned with predicted aromas associated with each label. (D) Aroma-object pairings are constructed. (E) The pairings are input into the proposed diffusion model to generate all near-neighboring molecules. (F) All molecules are correlated to the originating objects and stored in a list. (G) Engineer evaluates which objects and target compounds he/she wants to identify from the list and uses this information to down-select the number of olfaction sensors needed. (H) These sensors are integrated into the olfaction device of the robot. (I) Robot is ready for deployment.}
  \label{fig:coco-knn}
\end{figure}

Despite sensor advancements, the development of machine learning models for olfactory perception faces significant challenges, primarily due to the scarcity of comprehensive olfactory datasets and the difficulty in acquiring ground truth labels for the aromas associated with specific objects.
Humans describe chemical odourants with lingual descriptors such as \textit{fruity} and \textit{floral}, but sensors interpret odourants by their direct chemical reactions, leading to a disparity in data training methods for olfactory sensors (e.g. how does one describe to an olfactory sensor the "scent" of explosives?).
Datasets like those from Leffingwell \cite{leffingwell} and GoodScents \cite{goodscents} have been instrumental in providing foundational results.
However, due to the complexity in obtaining human-labeled odour descriptors over compounds \cite{lee2023}, these datasets are limited in scope, lacking complete coverage of the vast chemical space associated with olfactory stimuli \cite{sanchezlengeling2019machinelearningscentlearning} and the necessary human controls to ensure objective measurements such as compensation for genetic variation in olfactory mechanisms \cite{whitlock2020genetic}, environmental context \cite{mainland2014individual}, health status \cite{bratman_2024_olfaction_human_well-being, li2016olfactionparkinsons}, age \cite{doty1984influence}, and even potential olfaction system damage \cite{doty2001smellinjury}.

To address this data paucity, we propose a vision-olfaction-language dataset along with an innovative diffusion-based equivariant graph neural network (DEGNN) that allows one to minimize uncertainty whilst constructing olfactory datasets and selecting the appropriate sensors for an olfaction-equipped robot.
Our motivation is driven by the desire to teach robots to navigate by scent and triangulate the source of explosives, building on methods from \cite{burgues20-drone-chem-sense-survey, sniffybug-single-burgues19, sniffybug-swarm-duisterhof21, france2025_oio_method}.
Our research provides four primary contributions:
\begin{enumerate}
    \item Our model \textbf{aids in the selection of olfactory sensors} for a desired target compound or aroma.
    \item Our model increases the robustness of constructing olfactory datasets and \textbf{we provide an open dataset and training method for use by the community.}
    \item In result of (1) and (2), our model \textbf{reduces uncertainty due to the lack of grounding} in olfactory robotics tasks.
    \item In result of (3), our methods directly \textbf{increase accuracy and hardware efficiency} in scent-based navigation.
\end{enumerate}

We combine visual analysis, language processing, and molecular generation to enhance the field of olfactory machine learning through the construction of robust olfaction datasets.
By addressing the limitations of existing datasets and incorporating advanced sensor validation, we aim to advance the capabilities of artificial olfactory systems in accurately navigating by scent in complex environments.

\section{Background \& Related Work}

A significant hurdle in developing machine learning models for olfaction is the scarcity of extensive and high-quality datasets. 
Existing resources, such as the Leffingwell PMP 2001 \cite{leffingwell} and GoodScents \cite{goodscents} datasets, provide valuable information on various chemical compounds and their associated odour descriptors. 
However, these datasets are limited in the number of compounds they cover and do not encompass the full spectrum of olfactory stimuli. 
In addition, these datasets were constructed by human evaluators without all of the necessary controls needed to qualify the performance of their olfactory mechanisms leading to uncertainty in their grounding \cite{lee2023, sanchezlengeling2019machinelearningscentlearning}.
This is not a fault of the dataset authors, but a known problem in olfaction due to the limited range of human olfactory mechanisms \cite{whitlock2020genetic, mainland2014individual, bratman_2024_olfaction_human_well-being, li2016olfactionparkinsons, doty2001smellinjury}.
This limitation hampers the training of robust ML models capable of generalizing across diverse olfactory inputs, especially when many olfactory sensors exhibit better sensitivity than humans \cite{france-peddi-dennler-daescu-position_2025}.

\begin{figure*}[t!]
  \centering
  \includegraphics[width=0.8\linewidth]{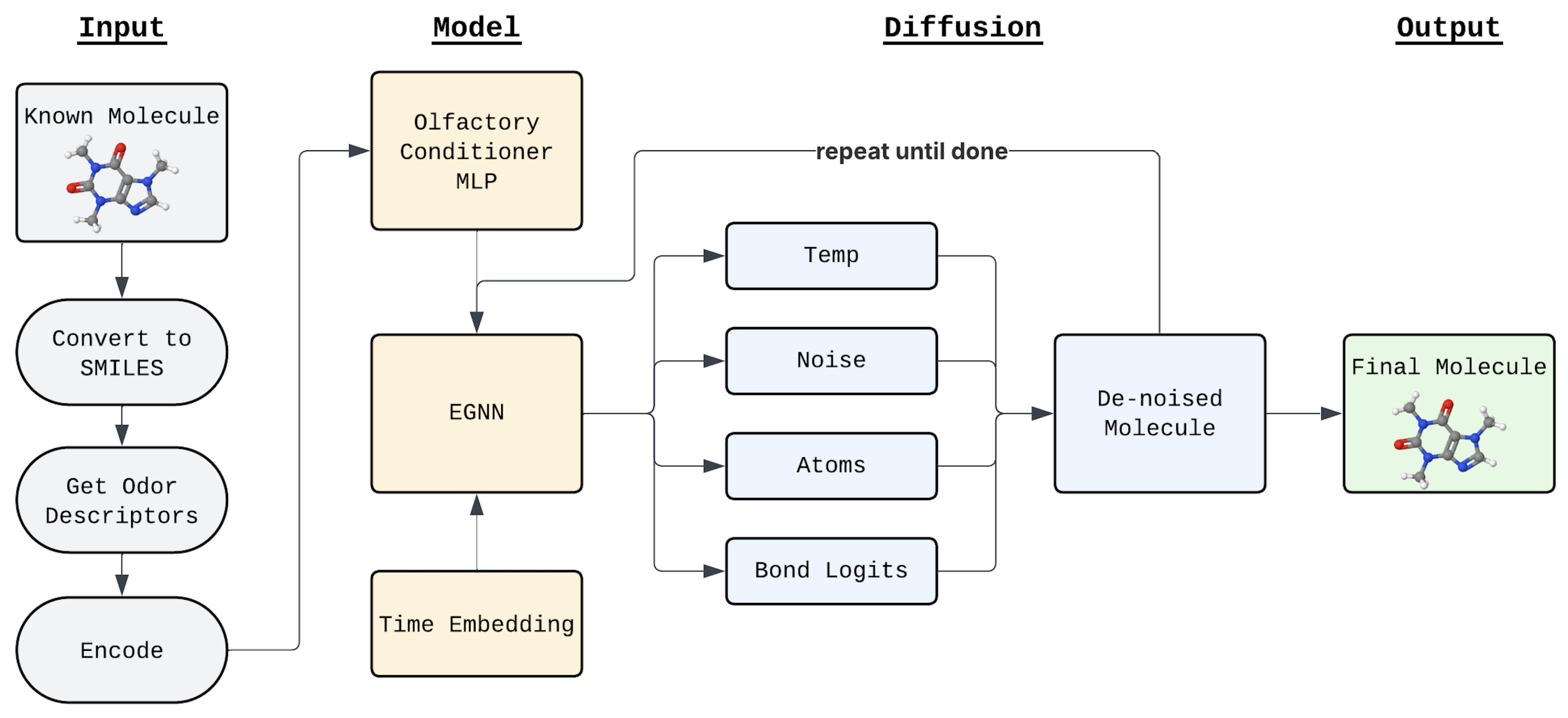}
  \caption{Overview of the olfactory-guided diffusion model architecture. Atom features and 3D coordinates evolve over time via EGNN layers, while bond types are predicted in parallel. The model is conditioned on the olfactory descriptor embeddings.}
  \label{fig:architecture}
\end{figure*}

To address the data scarcity and grounding issues, researchers have explored the use of vision-language models (VLMs) to generate odour descriptors from images. 
VLMs, trained on large-scale image-text pairs, can infer contextual information from visual inputs and generate corresponding textual descriptions \cite{radford2021clip} \cite{Wang2022-language-model-good-few-shot-learners}. 
By applying VLMs to images, it is possible to extract descriptive terms of objects within that may correlate to their olfactory characteristics. 
However, the reliability of these descriptors is contingent on the VLM's exposure to olfactory related data during training, the degree of which is unknown \cite{france-peddi-dennler-daescu-position_2025}. 
Consequently, the generated descriptors may lack specificity or accuracy in representing odours.

While efforts to construct olfaction-vision-language models (OVLMs) exist \cite{ovlm_france_app}, their underlying training data and training processes are not well standardized.
By analyzing visual content in a standard computer vision dataset like, for example, the COCO dataset \cite{cocodataset}, VLMs can infer potential olfactory characteristics associated with depicted scenes or objects. 
For example, if a motorcycle is identified in an image, the VLM will reason that carbon monoxide is likely a present molecule and will list the aromas that describe carbon monoxide given the appropriate prompt.
These inferred descriptors \textit{serve as a bridge} to map visual information to molecules, facilitating the identification of molecules likely present in a given image.

Thus, the reliance on VLMs introduces its own set of uncertainties, particularly regarding the accuracy and completeness of the generated odour descriptors which further contributes to the grounding problem.
To mitigate these concerns, we integrate a diffusion model trained on existing olfactory datasets and designed to generate novel molecular structures that correspond to the inferred odour descriptors, effectively expanding the chemical space beyond the limitations of current datasets. 
By doing so, we contribute to the oolfactory grounding problem by capturing a broader spectrum of odourants and molecules - including those not contained in the training data.

Our motivation in doing so is to obtain more reliable and accurate scent-based navigation in robots.
Sensors that rapidly sample the air only screen for a select set of molecules.
If a robot is given the task to navigate toward a specific odour, it has to be equipped with olfactory sensors that can target the specific molecules attributed to that odour.
By understanding the target molecule's near neighbors on the odour continuum, one can ensure to integrate all sensors that address these molecules as well.
As a simplified example, imagine that an unmanned aerial vehicle (UAV) is instructed to find the source of nitric oxide ($NO$), sometimes an indicator for the presence of explosives.
$NO$ can, in some environments, quickly oxidize to $NO_2$.
If the robot takes substantial time to navigate to the source of $NO$, it could be analyzing the wrong compound by missing the detection window, and at some point, may want to transition to detecting $NO_2$.
The olfactory sensors (for example, metal oxide sensors) required to detect both $NO$ and $NO_2$ are different, and it would be prudent for the engineer to query which near neighbor compounds could be relevant for detection so the UAV could be equipped with such sensors.
This issue is exemplified in tasks from \cite{burgues20-drone-chem-sense-survey} \cite{sniffybug-single-burgues19} \cite{sniffybug-swarm-duisterhof21} \cite{france2025_oio_method} \cite{france2025_oio_cal}.
We automate such queries in our dataset construction, such that a separate machine learning model is queried for the nearest neighbors of any compound identified by the VLM.
This not only informs better sensor selection, but also partially reconciles erroneous compounds produced by the VLM.

In machine olfaction, the Shape Theory of Olfaction suggests that molecules that exhibit similar structural traits "smell" similarly \cite{Billesbølle_2023_structure_human_odourant_receptor} \cite{Seshadri_2022_structure_why_does_that_molecule_smell} \cite{Wellawatte_2025_structure_xai} .
Although there is some support against this theory \cite{Block2015vto_refute} \cite{dyson1928vto} \cite{dyson1938vto} \cite{turin2015vto}, our proposed method assumes this theory is true, and we construct a model that builds off its principles.
For example, if a VLM indicates that the odour descriptors "fruity" and "floral" are associated with an object in an image, one can use the GoodScent or Leffingwell datasets to look up particular molecules that exude such aromas.
However, since these datasets are not comprehensive of all possible molecules, one can use the proposed model to generate "near neighbors" of molecules that are structurally similar to those found in the existing datasets.


To bridge the gap between odour descriptors and chemical compounds, diffusion models have emerged as a promising solution. 
Diffusion models are generative frameworks that learn to produce data samples by iteratively denoising random noise, guided by learned data distributions. 
In the context of molecular generation, diffusion models can be conditioned on textual inputs, such as odour descriptors, to generate novel molecular structures that correspond to the specified olfactory characteristics. 
This approach enables the exploration of chemical spaces beyond existing datasets, facilitating the discovery of new compounds with desired scent profiles. 
Recent studies have demonstrated the efficacy of text-guided diffusion models in generating molecules with specific properties, highlighting their potential in olfactory research  \cite{Luo2024-text-guided-molecule-diffusion} \cite{wang2025diffusionmodelsmoleculessurvey}.
Further research by Lee, et al. \cite{lee2023} and Sisson, et al. \cite{Sisson2025-gnn-for-aroma-chemical-bonds} have demonstrated feasibility of other generative models in producing viable molecules according to their desired odours or chemical bonds.

The validation of generated molecules necessitates empirical methods to confirm their olfactory properties. 
Advanced olfactory sensors offer a means to detect and analyze volatile compounds, providing data on their scent profiles. 
By comparing the sensor readings of generated molecules with the intended odour descriptors, researchers can assess the accuracy and relevance of the diffusion model's outputs. 
In reality, this empirical validation is very difficult for reasons denoted in the \textit{Limitations} section.
Work from Lee, et al. in \cite{lee2023} show how graph neural networks (GNNs) can be used to construct \textit{principal odour maps} for deducing probable aromas for a given molecule.
Our work here with GNNs strives to align with many of their assumptions, but leverages diffusion to overcome uncertainties associated with changing perceivable aromas among different concentrations of the same compound as exemplified by Longin, et al. \cite{longin_2020_bread_bias}.

Despite these advancements, challenges persist in accurately associating odours with their correct sources in complex environments. 
Our proposed methodology builds upon these existing works by integrating vision-language models to extract odour descriptors from images and employing diffusion models to generate corresponding molecular structures. 
We aim to enhance the robot's ability to disambiguate odour sources, improving navigation and decision-making in environments where olfactory cues are essential. 


\section{Methodology}
\subsection{Diffusion Model Selection}
Traditional methods for exploring chemical similarity, such as nearest neighbors using molecular fingerprints, are effective for retrieving known molecules from existing datasets. 
These methods are particularly well-suited for tasks that involve classification, clustering, or similarity-based search. 
However, they fall short when generating entirely new molecules that possess specific olfactory properties — especially combinations of scent descriptors not observed in known databases.

In contrast, diffusion models are generative frameworks capable of sampling novel molecular graphs from a learned distribution. When trained on molecular structures annotated with multi-label scent descriptors, diffusion models can conditionally generate new candidate molecules that reflect specified olfactory profiles (e.g., floral, musky, woody). This goes beyond simple retrieval — it enables creation and generation and thereby offers several key advantages:

Unlike k-nearest neighbor methods, diffusion models do not rely on existing database entries. They have generative capability and can synthesize new, chemically valid molecular structures that conform to desired scent characteristics.
Diffusion models also support multi-label conditioning, allowing fine control over the generated output. One can, for instance, request molecules that are simultaneously fruity, earthy, and sweet — even if no molecule with that combination exists in the training data.
By modeling a distribution over molecular structures, diffusion models allow for sampling of diverse outputs that satisfy the same olfactory constraints, improving the breadth of chemical space explored during generation.
The denoising process intrinsic to diffusion models implicitly learns smooth transitions in molecular structure and scent characteristics, enabling interpolation and optimization across scent manifolds.
The diffusion framework can naturally be extended to incorporate additional modalities such as synthetic accessibility, toxicity, volatility, or gas diffusive characteristics making it suitable for multi-objective molecule design.

While diffusion models require more computational resources to train and sample compared to nearest neighbors, their generative power and flexibility make them particularly well-suited for olfactory molecule discovery and scent-driven innovation. 
These properties are essential for developing novel fragrance molecules, identifying rare scent combinations, and automating early-stage formulation in perfumery, flavor chemistry, and environmental sensing.

\subsection{Graph Neural Networks}
Our use of equivariant graph neural networks (EGNNs) enhances our diffusion architecture.
EGNNs model atoms as both node features (e.g., atomic number), and continuous 3D positions. 
As a result, they learn from spatial relationships like distance and angle, and their outputs are equivariant to translation and rotation.
In turn, this enables geometry-aware bond type inference, better generalization to shape-driven scent effects, and modeling conformational flexibility, important for subscribing to the Shape Theory of Olfaction.

\section{Model Architecture}
\subsection{Diffusion Framework}

Our generative framework is built on a conditional denoising diffusion model, enhanced by an EGNN architecture which we denote as Diffusion EGNN, or DEGNN for brevity. 
This design is tailored to generate novel molecular structures in accordance with specific olfactory descriptors. Unlike traditional approaches that rely solely on molecular fingerprints or 2D-graph representations, our model incorporates 3D-molecular geometry and jointly learns both atomic identities and bond structures through time-reversible denoising steps.

Each molecule is represented as a graph \( G = (V, E) \), where nodes correspond to atoms and edges represent chemical bonds. We enrich this representation with 3D coordinates \( \mathbf{r}_i \in \mathbb{R}^3 \) for each atom, computed using RDKit's ETKDG embedding method \cite{Landrum2025-ib-rdkit}. Atoms are initially encoded as scalar atomic numbers, while bond types are annotated using categorical labels (single, double, triple, aromatic).

The model conditions generation on multi-label olfactory descriptors such as \textit{floral}, \textit{musky}, and \textit{fruity} on a separate \textit{olfactory conditioner}, which is a simple feedforward neural network. 
These descriptors are encoded as multi-hot binary vectors and projected into a continuous latent space via a learnable feedforward layer. This conditioning vector is then concatenated with both the node features and a time embedding to guide the diffusion process.

We adopt the standard forward diffusion process from a denoising diffusion probabilistic model (DDPM), where noise is incrementally added to the atomic node features over \( T \) timesteps. 
For a given molecule, the clean node features \( x_0 \) are perturbed into a noisy version \( x_t \) using:

\[
x_t = x_0 + \sqrt{\beta_t} \cdot \epsilon, \quad \epsilon \sim \mathcal{N}(0, I)
\]

where \( \beta_t \) is the variance schedule and \( t \in [1, T] \). We simplify this by scaling the noise linearly with \( t \).

The training objective is to predict the noise \( \epsilon \) added to \( x_0 \) given the noisy input \( x_t \), the timestep \( t \), and a conditioning vector representing the olfactory labels.

During training, Gaussian noise is progressively added to atomic features across discrete time steps \( t \in [1, T] \). 
In other words, the model is trained to reverse this process by predicting the original noise component \( \epsilon \) from the noisy input \( x_t \), conditioned on both time \( t \) and the olfactory label embedding. 
A simple linear time embedding module is employed to encode timestep information during the diffusion process.

Molecules in the GoodScents dataset are annotated with multi-label binary vectors \( y \in \{0, 1\}^L \), where \( L \) is the number of unique olfactory descriptors (e.g., floral, musky, fruity). These labels are projected into a latent conditioning space via a feedforward projection:

\[
\mathbf{c} = \text{Linear}(y)
\]

This conditioning vector \( \mathbf{c} \) is concatenated with both the node features and timestep encoding before being processed by the denoising model.
The core denoising mechanism comprises two stacked EGNN layers. Each EGNN layer performs equivariant message passing by computing distances and directional vectors between atoms, then applies the learned neural network to both update node features and apply geometry-aware coordinate shifts. Formally, for a node pair \( (i, j) \), the message \( m_{ij} \) is computed using:
\begin{equation}
    m_{ij} = \text{MLP}_\text{node}\left([x_i, x_j, \|\mathbf{r}_i - \mathbf{r}_j\|]\right)
\end{equation}

A corresponding coordinate update is also computed:
\begin{equation}
    \Delta \mathbf{r}_i = \sum_{j \in \mathcal{N}(i)} \text{MLP}_\text{coord}(\|\mathbf{r}_i - \mathbf{r}_j\|) \cdot (\mathbf{r}_i - \mathbf{r}_j)
\end{equation}

These updates preserve equivariance under rigid-body transformations, allowing the model to respect the geometric nature of molecular structures. The aggregated node messages are summed and added to the current feature representation \( x_i \), while the position updates are added to \( \mathbf{r}_i \).

In addition to denoising atomic identities, the model includes a parallel bond type predictor. For each edge in the graph, it concatenates the embeddings of the two endpoint nodes and predicts a bond type using a four-way softmax classifier. This classifier is supervised via a cross-entropy loss computed against ground truth bond labels.

The training objective loss $\mathcal{L}_{\text{total}}$ thus consists of two components: a mean squared error (MSE) loss $\mathcal{L}_{\text{MSE}}$ for node denoising, and a cross-entropy loss $\mathcal{L}_{\text{CE}}$ for bond classification:
\begin{equation}
\mathcal{L}_{\text{total}} = \mathcal{L}_{\text{MSE}}(x_t, \hat{x}_0) + \mathcal{L}_{\text{CE}}
\end{equation}

Once trained, generation begins from Gaussian noise. A noisy vector \( x_T \) and random 3D coordinates are initialized and iteratively denoised over \( T \) steps. After the final step, atomic predictions are rounded and filtered to chemically reasonable elements (e.g., C, N, O, S, Cl). The model also predicts bond types, and a molecular graph is assembled and sanitized using RDKit. If valid, the final structure is converted to a canonical SMILES string and visualized.
Conditioning vectors can be customized (e.g., setting `floral` and `fruity` to 1, others to 0) to generate novel molecules with targeted scent profiles.

To enforce chemical validity, we include filtering rules during decoding. Atoms with unstable valence configurations or rare atomic numbers are discarded. Additionally, we apply \texttt{nan\_to\_num} to model outputs to prevent numerical instabilities. Molecules are validated and visualized using RDKit's sanitization and rendering tools.
During training, we apply temperature-based softmax to the bond logits $\sigma_{bond}$ according to the following:

\begin{equation}
\sigma_{scaled} = \sigma_{bond} / \tau
\end{equation}

\noindent Where $\tau$ represents temperature.
This helps reduce overly random predictions and improve structural validity.
Figure~\ref{fig:architecture} illustrates the generation process defined above.

In summary, our model provides a 3D-aware, scent-conditioned molecular generator grounded in diffusion theory and equivariant geometry. 
It enables controlled exploration of molecular space according to sensory targets and opens a new pathway for olfactory-driven molecular design.

\begin{figure*}[t!]
  \centering
  \includegraphics[width=138mm]{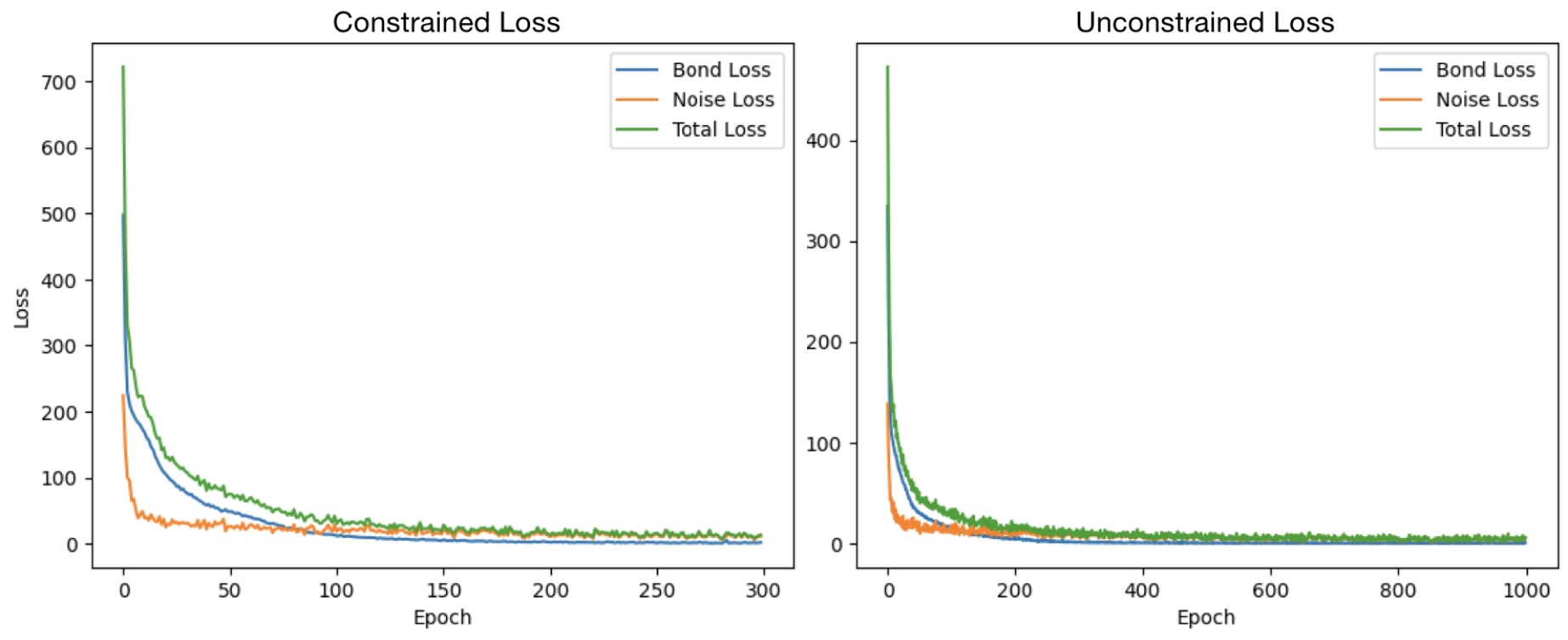}
  \caption{Loss curves of the final trained models for both the constrained and unconstrained variants.}
  \label{fig:diffuseloss}
\end{figure*}


\subsection{Molecular Validation}

Our diffusion model needs a series of checks and balances to ensure no nonsensical molecules are produced.
To do this, we incorporate a series of validation steps into the generation.

Firstly, we validate the atomic number range.
The periodic table only defines elements up to atomic number 118. 
Diffusion models might predict out-of-range values (e.g., negative or exaggerated atomic numbers) due to noise. 
Skipping such atoms ensures that the generated molecule remains within the bounds of known chemistry.
This validation prevents the addition of chemically nonsensical atoms to the molecule. 

We also ensure edge deduplication by avoiding the creation of multiple redundant bonds between the same pair of atoms.
Molecular graphs are undirected and generally allow only one bond per atom pair (with varying types). 
This check ensures graph realism and prevents downstream errors in RDKit, which will reject duplicate edges unless explicitly defined as aromatic or resonance structures.

To validate the bond type, we add a heuristic to bond type inference.
This validation converts continuous-valued output of the diffusion model into discrete bond types (single, double, triple).
Since the model doesn’t explicitly predict bond types, this heuristic uses the difference in atomic number features as a proxy. 
While not chemically perfect, it introduces structural diversity and prevents over-simplified molecules (e.g., with only single bonds).
Our model also reconciles bonding errors that may occur during inference.
Even after edge deduplication and bond-type inference, some atom pairs may still be chemically incompatible for bonding (e.g., noble gases or already saturated atoms). This try/except block avoids crashes and skips invalid bond additions.

Other sanitization techniques of our model involve the addition of implicit hydrogens,  validation of molecular valence and connectivity, and generation of correct kekulé or aromatic representations.
Sanitization is critical for ensuring that the generated molecule is chemically plausible and renderable, especially for descriptor computation or visualization.
These sanitization techniques check and correct the molecular structure using RDKit’s built-in rules.

Even with the above rules in place, invalid molecules may still result from our diffusion model.
Molecules that pass validation up to this point are further checked to assess unbalanced aromatic rings or invalid formal charges.
This acts as a final safeguard before converting the final molecule to its SMILES and prevents flawed outputs from leaking into evaluation metrics or dataset augmentation.
After SMILES conversion, we perform one final check to ensure human-readability, existence of the SMILES against a known database, and compatibility with downstream chemoinformatics tools.

\section{Experiments \& Results}

We first train the EGNN and diffusion model on the LeffingWell and GoodScent datasets.
We leverage the dataset from \cite{lee2023} as it succinctly combines both datasets into a methodical format. 
The resulting dataset is nearly 5000 samples in size.
While this is a small dataset in comparison to modern machine learning tasks in other modalities like image recognition, it is rare to have a dataset this size in the realm of olfaction.
We use an 80-20 train-test split in which we evaluate our models over the indicated data.
We find that diffusing between 800-1200 steps is ideal, and perform our experiments at 1000 steps.
We train the EGNN for 1000 epochs with an embedding dimension of size 8.
We train two models in the exact same manner with the exception that one model is unconstrained in that it can generate molecules from any available atom; the other model is "constrained" in that it can only generate molecules from the following atoms: \textit{C, N, O, F, P, S,} and \textit{Cl}.
The training loss for both models is shown in Figure \ref{fig:diffuseloss}.
The final results are summarized in Table \ref{tab:table1}.

We expected to see better performance by constraining the diffusion to a specific set of atoms.
However, this actually resulted in worse performance, generating validated molecules over less than 10\% of samples. 
We suspect this is because the permutation of such a small set of elements did not provide a diverse enough dataset from which the diffusion model could generate neighboring compounds in most cases.
While the unconstrained model is more complex, we find that the increased sample space in combination with a longer training time yielded better performance than the constrained model and we leverage that model to communicate our final results.

Although molecules are generated for every set of descriptors that are input to the model, we find that only 27.71\% of molecules diffused from our test set pass our molecular validation checks. 
This is expected and welcome as we suspect not every permutation of identified odour descriptors returns a large array of near-neighbor molecules.
Note that the GoodScents and LeffingWell datasets that we leverage for training data do not include \textit{all} feasible compounds and odours. \footnote{For more information on the limitations of the Leffingwell and GoodScent datasets, please consult \cite{leffingwell} and \cite{goodscents}, respectively. 
We acknowledge that our approach here inherits any limitations noted in these datasets, including those attributed to the subjectivity of human-produced olfactory labels. However, we hope that our diffusion model partially reconciles some of the uncertainty associated with these olfactory labels.}

We then test our proposed model with the VLM over the full test data of the COCO dataset \cite{cocodataset}. 
We select GPT-4o \cite{openai2024gpt4ocard} as our VLM and prompt it to caption each image with odour descriptors that are suspected to be present.
An array of odour descriptors is then associated with each image in the test set and input into our diffusion-EGNN model.
We obtain more permutations of odour descriptors than what are available in our training data, which leads to more possible molecules being produced than those experienced in our training set.
Because of this, we note a 28.20\% success rate in diffusing near-neighbor molecules from VLM odour descriptors after validation checks are passed.
This marks a pleasing transfer of training accuracy to test accuracy and amounts evidence for us to suggest that the aromas for the sample of molecules provided in the consolidated Leffingwell and GoodScent datasets may be good proxies of the much larger molecular distribution.

\subsection{Additive vs Subtractive Cases}
The DEGNN method can improve hardware designs in two ways: one where it advises sensors to be \textit{added} to the olfaction device in order to improve better coverage of the target odours and one where it advises to \textit{subtract} sensors from the device.

\begin{figure}[t!]
  \centering
  \includegraphics[width=0.7\linewidth]{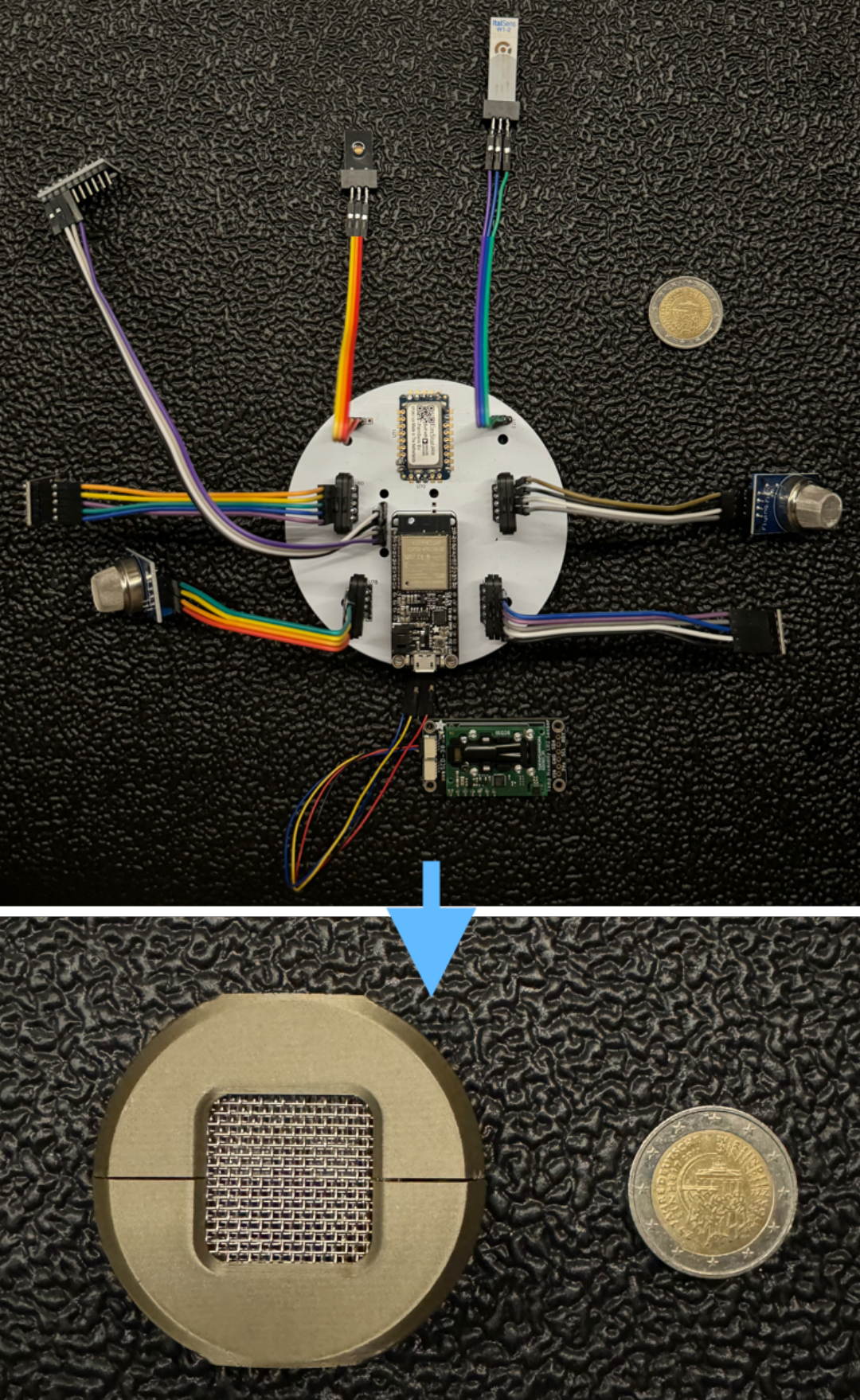}
  \caption{An example where DEGNN helped reduce the number of sensors from 16 sensors (top) to 4 sensors (bottom) for a robot tracking ammonia, reducing the size, weight, power, and compute needed for the olfaction device. A coin is placed next to each to establish scale.}
  \label{fig:breakout}
\end{figure}

In most scenarios, DEGNN can advise \textit{which} sensors to add to the olfaction device.
However, DEGNN can eliminate over-engineering of the olfaction device by reducing the number of sensors required to capture a target odour. 
For example, in one case study shown in Figure \ref{fig:breakout}, DEGNN enabled the reduction from 16 sensors to 4 sensors. 
This confirmation and the ability to relieve sensors helps save weight, power, and compute aboard the robot.

\vspace{-1mm}
\begin{table}[htbp]
\caption{Final results: Percentage of test samples with chemically-viable generated molecules.}
\begin{center}
\begin{tabular}{|c|c|c|}
\hline
\textbf{Dataset}&
\textbf{Constrained}& \textbf{Unconstrained} \\
\hline
GS \& LW Test Set \cite{lee2023}& $< 10\%$ & 27.71\%\\
COCO Test Set \cite{cocodataset}& $< 10\%$ & 28.20\% \\
\hline
\end{tabular}
\label{tab:table1}
\end{center}
\end{table}

\vspace{-1mm}

\section{Limitations}
\vspace{-1mm}

While our integrated framework combining VLMs and diffusion-based molecular generation offers a novel approach to enhancing robotic odour source localization, it is essential to acknowledge its limitations and the challenges that persist in scent-based navigation.
VLMs, though powerful in bridging visual and textual modalities, are not specifically trained on olfactory datasets. 
Consequently, their ability to generate accurate and comprehensive odour descriptors from images is constrained. 
This limitation can lead to incomplete or imprecise odour representations, affecting the grounding of subsequent molecular generation process. 
Moreover, VLMs may struggle with contextual reasoning and spatial understanding, which are crucial for accurately associating odours with their sources in complex environments.
This can be analogously observed from the work of Xie, et al. in \cite{xie2024sonicvisionlm} where they attempt to infer sound from images using VLMs.
Our dataset and DEGNN method here inherit these limitations as a consequence.
For example, we noticed in our training that the VLM tends to associate carbon monoxide to the presence of a vehicle in the COCO image, but does not consider the fact that the vehicle may be electric.
However, due to the lack of robotics-centric olfaction datasets, we emphasize that our methodology holds value.

The diffusion model, trained on existing olfactory datasets, aims to generate novel molecular structures corresponding to inferred odour descriptors.
However, the quality and diversity of the training data directly influence the model's generative capabilities. 
Given the limited scope of current olfactory datasets, the diffusion model may not capture the full spectrum of possible odourants or molecules, potentially leading to gaps in odour representation. 
This is exemplified in the fact that we observed more valid molecules generated with the VLM data than with the training set.

Additionally, our method gives heavy credence to the Shape Theory of Olfaction.
If this theory is proven untrue, it may invalidate the efficacy of our method.
The generated molecules require empirical validation to confirm their olfactory properties which can be accomplished through various olfaction sensors.
However, this can be very resource-intensive as the instrumentation required to validate the presence of compounds is expensive.
We note that Lee et, al. \cite{lee2023} also observed similar attributes whose work of which we build on top.
In addition, obtaining all possible molecules over which to evaluate said sensors can be restrictive due to regulations and required licenses.
Finally, even if one could obtain testing samples of all possible compounds in a unanimous quantity, it is not enough to test each compound individually.
The combination and interaction of certain compounds produce entirely new odour descriptors which are not yet entirely predictable.
Rapidly quantifying the presence of compounds within an air sample and the aromas attributed to them is a known problem within olfaction; this underscores the need for more datasets, learning techniques, and grounding methods as proposed here.

Another significant challenge in olfactory navigation is the accurate association of detected odours with their correct sources. 
Environmental factors such as airflow dynamics, presence of multiple odour sources, and obstacles can cause odour plumes to disperse unpredictably, leading to potential misattribution of odours to incorrect objects. 
While our framework enhances the robot's ability to infer and generate potential odourant molecules, it does not eliminate the possibility of such misassociations. 
Therefore, our system may still see difficulties in environments with complex odour landscapes.

Implementing the proposed framework in real-time robotic systems poses computational challenges. 
The integration of VLMs, diffusion models, and olfactory sensors requires efficient processing capabilities to ensure timely decision-making during navigation. 
Latency in processing can hinder the robot's responsiveness, especially in dynamic environments where rapid adaptation is necessary.
It should be noted that the proposed method is intended to aid in the selection of which olfactory sensors should be integrated on a robot to navigate to a particular compound, prior to deployment.

In summary of the above, we acknowledge that there are inherent limitations of our proposed methodology, but hope that it can be used to generate highly probable compounds for given aromas when constructing vision-olfactory datasets and informing sensor selection in olfactory robotics tasks.

\vspace{-1mm}
\section{Conclusion}
\vspace{-1mm}
Machine olfaction is still a young area of AI and robotics that receives disproportionate attention and standardization in comparison to other modalities such as computer vision and audition.
This creates several opportunities for dataset construction, standardization, and benchmarking.
More attention is especially needed to adapt these methods to complex environments such as robotic OSL to toxic compounds.
Our methodology represents a foundational step towards improving sensor selection in olfactory robotics by integrating visual, linguistic, and olfactory data. 
We note opportunities to utilize our method as both a standalone algorithm and as part of an automation pipeline used to construct olfactory datasets via VLMs.
However, we acknowledge it is as a foundational method upon which can be improved by the community. 
The limitations outlined above highlight the opportunities for continued research to address the complexities inherent in olfactory perception. 
Our method enhances a robot's ability to navigate by scent and reconciles known shortcomings in both the construction of olfaction-vision datasets, but we hope it inspires more researchers in the field to enter olfactory robotics and contribute to solving the olfactory grounding problem.


\bibliography{sample-base}

\begin{thebibliography}{10}
\providecommand{\url}[1]{#1}
\csname url@samestyle\endcsname
\providecommand{\newblock}{\relax}
\providecommand{\bibinfo}[2]{#2}
\providecommand{\BIBentrySTDinterwordspacing}{\spaceskip=0pt\relax}
\providecommand{\BIBentryALTinterwordstretchfactor}{4}
\providecommand{\BIBentryALTinterwordspacing}{\spaceskip=\fontdimen2\font plus
\BIBentryALTinterwordstretchfactor\fontdimen3\font minus \fontdimen4\font\relax}
\providecommand{\BIBforeignlanguage}[2]{{%
\expandafter\ifx\csname l@#1\endcsname\relax
\typeout{** WARNING: IEEEtran.bst: No hyphenation pattern has been}%
\typeout{** loaded for the language `#1'. Using the pattern for}%
\typeout{** the default language instead.}%
\else
\language=\csname l@#1\endcsname
\fi
#2}}
\providecommand{\BIBdecl}{\relax}
\BIBdecl

\bibitem{Chowdhury2025_survey}
\BIBentryALTinterwordspacing
M.~A.~Z. Chowdhury and M.~A. Oehlschlaeger, ``Artificial intelligence in gas sensing: A review,'' \emph{ACS Sensors}, vol.~10, no.~3, pp. 1538--1563, Mar 2025. [Online]. Available: \url{https://doi.org/10.1021/acssensors.4c02272}
\BIBentrySTDinterwordspacing

\bibitem{covington2021_artificialolfactionsurvey21stcentury}
J.~Covington, S.~Marco, K.~Persaud, S.~Schiffman, and H.~Nagle, ``\BIBforeignlanguage{English}{Artificial olfaction in the 21st century},'' \emph{\BIBforeignlanguage{English}{IEEE Sensors Journal}}, vol.~21, no.~11, pp. 12\,969--12\,990, Jun. 2021.

\bibitem{gutierrez_osuna_machine_olfaction_survey_2002}
R.~Gutierrez-Osuna, ``Pattern analysis for machine olfaction: a review,'' \emph{IEEE Sensors Journal}, vol.~2, no.~3, pp. 189--202, 2002.

\bibitem{kim_olfactory_sensor_survey_2022}
\BIBentryALTinterwordspacing
C.~Kim, K.~K. Lee, M.~S. Kang, D.-M. Shin, J.-W. Oh, C.-S. Lee, and D.-W. Han, ``Artificial olfactory sensor technology that mimics the olfactory mechanism: a comprehensive review,'' \emph{Biomaterials Research}, vol.~26, no.~1, p.~40, 2022. [Online]. Available: \url{https://spj.science.org/doi/abs/10.1186/s40824-022-00287-1}
\BIBentrySTDinterwordspacing

\bibitem{leffingwell}
L.~. Associates, ``Pmp 2001 - database of perfumery materials and performance,'' \url{http://www.leffingwell.com/bacispmp.htm}, accessed: 2025-03-08.

\bibitem{goodscents}
T.~G.~S. Company, ``The good scents company information system,'' \url{http://www.thegoodscentscompany.com/}, accessed: 2025-03-08.

\bibitem{lee2023}
\BIBentryALTinterwordspacing
B.~K. Lee, E.~J. Mayhew, B.~Sanchez-Lengeling, J.~N. Wei, W.~W. Qian, K.~A. Little, M.~Andres, B.~B. Nguyen, T.~Moloy, J.~Yasonik, J.~K. Parker, R.~C. Gerkin, J.~D. Mainland, and A.~B. Wiltschko, ``A principal odor map unifies diverse tasks in olfactory perception,'' \emph{Science}, vol. 381, no. 6661, pp. 999--1006, 2023. [Online]. Available: \url{https://www.science.org/doi/abs/10.1126/science.ade4401}
\BIBentrySTDinterwordspacing

\bibitem{sanchezlengeling2019machinelearningscentlearning}
\BIBentryALTinterwordspacing
B.~Sanchez-Lengeling, J.~N. Wei, B.~K. Lee, R.~C. Gerkin, A.~Aspuru-Guzik, and A.~B. Wiltschko, ``Machine learning for scent: Learning generalizable perceptual representations of small molecules,'' 2019. [Online]. Available: \url{https://arxiv.org/abs/1910.10685}
\BIBentrySTDinterwordspacing

\bibitem{whitlock2020genetic}
K.~E. Whitlock, ``Genetic variation in human odor perception,'' \emph{Chemical Senses}, vol.~45, pp. 549--556, 2020.

\bibitem{mainland2014individual}
J.~D. Mainland, A.~Keller, Y.~Li, T.~Zhou, C.~Trimmer, L.~L. Snyder, A.~H. Moberly, K.~A. Adipietro, W.~L. Liu, H.~Zhuang \emph{et~al.}, ``The missense of smell: functional variability in the human odorant receptor repertoire,'' \emph{Nature Neuroscience}, vol.~17, no.~1, pp. 114--120, 2014.

\bibitem{bratman_2024_olfaction_human_well-being}
\BIBentryALTinterwordspacing
G.~N. Bratman, C.~Bembibre, G.~C. Daily, R.~L. Doty, T.~Hummel, L.~F. Jacobs, P.~H. Kahn, C.~Lashus, A.~Majid, J.~D. Miller, A.~Oleszkiewicz, H.~Olvera-Alvarez, V.~Parma, A.~M. Riederer, N.~L. Sieber, J.~Williams, J.~Xiao, C.-P. Yu, and J.~D. Spengler, ``Nature and human well-being: The olfactory pathway,'' \emph{Science Advances}, vol.~10, no.~20, p. eadn3028, 2024. [Online]. Available: \url{https://www.science.org/doi/abs/10.1126/sciadv.adn3028}
\BIBentrySTDinterwordspacing

\bibitem{li2016olfactionparkinsons}
C.~Li, K.~Nie, Y.~Ma, M.~Shao, and G.~Li, ``Olfactory dysfunction in parkinson’s disease: a systematic review and meta-analysis,'' \emph{Frontiers in Aging Neuroscience}, vol.~8, p. 145, 2016.

\bibitem{doty1984influence}
R.~L. Doty, P.~Shaman, S.~L. Applebaum, R.~Giberson, L.~Siksorski, and L.~Rosenberg, ``Influence of age and age-related diseases on olfactory function,'' \emph{Annals of the New York Academy of Sciences}, vol. 435, no.~1, pp. 36--49, 1984.

\bibitem{doty2001smellinjury}
R.~L. Doty and V.~Kamath, ``Smell identification ability: changes with age,'' \emph{Science of Aging Knowledge Environment}, vol. 2001, no.~10, p. cp2, 2001.

\bibitem{burgues20-drone-chem-sense-survey}
\BIBentryALTinterwordspacing
J.~Burgués and S.~Marco, ``Environmental chemical sensing using small drones: A review,'' \emph{Science of The Total Environment}, vol. 748, p. 141172, 2020. [Online]. Available: \url{https://www.sciencedirect.com/science/article/pii/S004896972034701X}
\BIBentrySTDinterwordspacing

\bibitem{sniffybug-single-burgues19}
\BIBentryALTinterwordspacing
J.~Burgués, V.~Hernández, A.~J. Lilienthal, and S.~Marco, ``Smelling nano aerial vehicle for gas source localization and mapping,'' \emph{Sensors}, vol.~19, no.~3, 2019. [Online]. Available: \url{https://www.mdpi.com/1424-8220/19/3/478}
\BIBentrySTDinterwordspacing

\bibitem{sniffybug-swarm-duisterhof21}
B.~P. Duisterhof, S.~Li, J.~Burgués, V.~J. Reddi, and G.~C. H.~E. de~Croon, ``Sniffy bug: A fully autonomous swarm of gas-seeking nano quadcopters in cluttered environments,'' in \emph{2021 IEEE/RSJ International Conference on Intelligent Robots and Systems (IROS)}, 2021, pp. 9099--9106.

\bibitem{france2025_oio_method}
\BIBentryALTinterwordspacing
K.~K. France and O.~Daescu, ``Olfactory inertial odometry: Methodology for effective robot navigation by scent,'' 2025. [Online]. Available: \url{https://arxiv.org/abs/2506.02373}
\BIBentrySTDinterwordspacing

\bibitem{france-peddi-dennler-daescu-position_2025}
\BIBentryALTinterwordspacing
K.~K. France, R.~Peddi, N.~Dennler, and O.~Daescu, ``Position: Olfaction standardization is essential for the advancement of embodied artificial intelligence,'' 2025. [Online]. Available: \url{https://arxiv.org/abs/2506.00398}
\BIBentrySTDinterwordspacing

\bibitem{radford2021clip}
\BIBentryALTinterwordspacing
A.~Radford, J.~W. Kim, C.~Hallacy, A.~Ramesh, G.~Goh, S.~Agarwal, G.~Sastry, A.~Askell, P.~Mishkin, J.~Clark, G.~Krueger, and I.~Sutskever, ``Learning transferable visual models from natural language supervision,'' 2021. [Online]. Available: \url{https://arxiv.org/abs/2103.00020}
\BIBentrySTDinterwordspacing

\bibitem{Wang2022-language-model-good-few-shot-learners}
Z.~Wang, M.~Li, R.~Xu, L.~Zhou, J.~Lei, X.~Lin, S.~Wang, Z.~Yang, C.~Zhu, D.~Hoiem, S.-F. Chang, M.~Bansal, and H.~Ji, ``Language models with image descriptors are strong few-shot video-language learners,'' in \emph{Proceedings of the 36th International Conference on Neural Information Processing Systems}, ser. NIPS '22.\hskip 1em plus 0.5em minus 0.4em\relax Red Hook, NY, USA: Curran Associates Inc., 2022.

\bibitem{ovlm_france_app}
\BIBentryALTinterwordspacing
K.~K. France, ``Scentience,'' \url{https://apps.apple.com/us/app/scentience/id6741092923}, 2025, accessed: 2025-03-02. [Online]. Available: \url{https://apps.apple.com/us/app/scentience/id6741092923}
\BIBentrySTDinterwordspacing

\bibitem{cocodataset}
\BIBentryALTinterwordspacing
T.~Lin, M.~Maire, S.~J. Belongie, L.~D. Bourdev, R.~B. Girshick, J.~Hays, P.~Perona, D.~Ramanan, P.~Doll{'{a} }r, and C.~L. Zitnick, ``Microsoft {COCO:} common objects in context,'' \emph{CoRR}, vol. abs/1405.0312, 2014. [Online]. Available: \url{http://arxiv.org/abs/1405.0312}
\BIBentrySTDinterwordspacing

\bibitem{france2025_oio_cal}
\BIBentryALTinterwordspacing
K.~K. France, O.~Daescu, A.~Paul, and S.~Prasad, ``Olfactory inertial odometry: Sensor calibration and drift compensation,'' 2025. [Online]. Available: \url{https://arxiv.org/abs/2506.04539}
\BIBentrySTDinterwordspacing

\bibitem{Billesbølle_2023_structure_human_odourant_receptor}
\BIBentryALTinterwordspacing
C.~B. Billesb{\o}lle, C.~A. de~March, W.~J.~C. van~der Velden, N.~Ma, J.~Tewari, C.~L. del Torrent, L.~Li, B.~Faust, N.~Vaidehi, H.~Matsunami, and A.~Manglik, ``Structural basis of odorant recognition by a human odorant receptor,'' \emph{Nature}, vol. 615, no. 7953, pp. 742--749, Mar 2023. [Online]. Available: \url{https://doi.org/10.1038/s41586-023-05798-y}
\BIBentrySTDinterwordspacing

\bibitem{Seshadri_2022_structure_why_does_that_molecule_smell}
A.~Seshadri, H.~A. Gandhi, G.~P. Wellawatte, and A.~D. White, ``Why does that molecule smell?'' Dec. 2022, chemRxiv.

\bibitem{Wellawatte_2025_structure_xai}
\BIBentryALTinterwordspacing
G.~P. Wellawatte and P.~Schwaller, ``Human interpretable structure-property relationships in chemistry using explainable machine learning and large language models,'' \emph{Communications Chemistry}, vol.~8, no.~1, p.~11, Jan 2025. [Online]. Available: \url{https://doi.org/10.1038/s42004-024-01393-y}
\BIBentrySTDinterwordspacing

\bibitem{Block2015vto_refute}
E.~Block, S.~Jang, H.~Matsunami, S.~Sekharan, B.~Dethier, M.~Z. Ertem, S.~Gundala, Y.~Pan, S.~Li, Z.~Li, S.~N. Lodge, M.~Ozbil, H.~Jiang, S.~F. Penalba, V.~S. Batista, and H.~Zhuang, ``\BIBforeignlanguage{eng}{Implausibility of the vibrational theory of olfaction},'' \emph{\BIBforeignlanguage{eng}{Proceedings of the National Academy of Sciences of the United States of America}}, vol. 112, no.~21, pp. E2766--E2774, May 2015, research Support, N.I.H., Extramural; Research Support, Non-U.S. Gov't; Research Support, U.S. Gov't, Non-P.H.S.

\bibitem{dyson1928vto}
G.~Dyson, ``Some aspects of the vibration theory of odor,'' \emph{Perfumery and essential oil record}, vol.~19, no. 456-459, 1928.

\bibitem{dyson1938vto}
\BIBentryALTinterwordspacing
G.~Malcolm~Dyson, ``The scientific basis of odour,'' \emph{Journal of the Society of Chemical Industry}, vol.~57, no.~28, pp. 647--651, 1938. [Online]. Available: \url{https://onlinelibrary.wiley.com/doi/abs/10.1002/jctb.5000572802}
\BIBentrySTDinterwordspacing

\bibitem{turin2015vto}
\BIBentryALTinterwordspacing
L.~Turin, S.~Gane, D.~Georganakis, K.~Maniati, and E.~M.~C. Skoulakis, ``Plausibility of the vibrational theory of olfaction,'' \emph{Proceedings of the National Academy of Sciences}, vol. 112, no.~25, pp. E3154--E3154, 2015. [Online]. Available: \url{https://www.pnas.org/doi/abs/10.1073/pnas.1508035112}
\BIBentrySTDinterwordspacing

\bibitem{Luo2024-text-guided-molecule-diffusion}
Y.~Luo, J.~Fang, S.~Li, Z.~Liu, J.~Wu, A.~Zhang, W.~Du, and X.~Wang, ``\BIBforeignlanguage{en}{Text-guided small molecule generation via diffusion model},'' \emph{\BIBforeignlanguage{en}{iScience}}, vol.~27, no.~11, p. 110992, Sep. 2024.

\bibitem{wang2025diffusionmodelsmoleculessurvey}
\BIBentryALTinterwordspacing
L.~Wang, C.~Song, Z.~Liu, Y.~Rong, Q.~Liu, S.~Wu, and L.~Wang, ``Diffusion models for molecules: A survey of methods and tasks,'' 2025. [Online]. Available: \url{https://arxiv.org/abs/2502.09511}
\BIBentrySTDinterwordspacing

\bibitem{Sisson2025-gnn-for-aroma-chemical-bonds}
L.~Sisson, A.~A. Barsainyan, M.~Sharma, and R.~Kumar, ``\BIBforeignlanguage{en}{Deep learning for odor prediction on aroma-chemical blends},'' \emph{\BIBforeignlanguage{en}{ACS Omega}}, vol.~10, no.~9, pp. 8980--8992, Mar. 2025.

\bibitem{longin_2020_bread_bias}
\BIBentryALTinterwordspacing
F.~Longin, H.~Beck, H.~Gütler, W.~Heilig, M.~Kleinert, M.~Rapp, N.~Philipp, A.~Erban, D.~Brilhaus, T.~Mettler-Altmann, and B.~Stich, ``Aroma and quality of breads baked from old and modern wheat varieties and their prediction from genomic and flour-based metabolite profiles,'' \emph{Food Research International}, vol. 129, p. 108748, 2020. [Online]. Available: \url{https://www.sciencedirect.com/science/article/pii/S0963996919306349}
\BIBentrySTDinterwordspacing

\bibitem{Landrum2025-ib-rdkit}
G.~Landrum, P.~Tosco, B.~Kelley, R.~Rodriguez, D.~Cosgrove, R.~Vianello, {sriniker}, P.~Gedeck, G.~Jones, E.~Kawashima, {NadineSchneider}, D.~Nealschneider, A.~Dalke, M.~Swain, B.~Cole, S.~Turk, A.~Savelev, {tadhurst-cdd}, A.~Vaucher, M.~W{\'o}jcikowski, I.~Take, R.~Walker, V.~F. Scalfani, H.~Faara, K.~Ujihara, D.~Probst, J.~Lehtivarjo, G.~Godin, A.~Pahl, and N.~Maeder, ``rdkit/rdkit: 2025\_03\_2 (q1 2025) release,'' 2025.

\bibitem{openai2024gpt4ocard}
\BIBentryALTinterwordspacing
O.~and:, A.~Hurst, A.~Lerer, and e.~a. Adam P.~Goucher, ``Gpt-4o system card,'' 2024. [Online]. Available: \url{https://arxiv.org/abs/2410.21276}
\BIBentrySTDinterwordspacing

\bibitem{xie2024sonicvisionlm}
Z.~Xie, S.~Yu, M.~Li, Q.~He, C.~Chen, and Y.-G. Jiang, ``Sonicvisionlm: Playing sound with vision language models,'' \emph{arXiv preprint arXiv:2401.04394}, 2024.

\end{thebibliography}
\bibliographystyle{IEEEtran}

\end{document}